\documentclass[10pt,twocolumn,letterpaper]{article}

\usepackage{iccv}
\usepackage{times}
\usepackage{epsfig}
\usepackage{graphicx}
\usepackage{amsmath}
\usepackage{amssymb}
\usepackage{bm}
\usepackage{url}
\usepackage{subcaption}
\usepackage{xcolor}
\usepackage{lipsum}

\usepackage[pagebackref=true,breaklinks=true,letterpaper=true,colorlinks,bookmarks=false]{hyperref}

\iccvfinalcopy 

\def\iccvPaperID{10} 

\ificcvfinal\fi

\newcommand{\figref}[1]{Figure~\ref{fig:#1}}
\newcommand{\secref}[1]{Section~\ref{sec:#1}}

\newcommand{\cdms}{\,cd/m$^2$\xspace}

\LetLtxMacro{\originaleqref}{\eqref}
\renewcommand{\eqref}[1]{Eq.~\originaleqref{eq:#1}}

\newcommand{\pgraph}[1]{\vspace{1mm}\noindent\textbf{#1}:}

\begin{document}


\title{How to cheat with metrics in single-image HDR reconstruction}

\author{Gabriel Eilertsen$^1$,\; Saghi Hajisharif$^1$,\; Param Hanji$^2$,\; Apostolia Tsirikoglou$^1$,\\ Rafa\l~K. Mantiuk$^2$,\; Jonas Unger$^1$
  \vspace{0.07cm}\\
  $^1$~Dept. of Science and Technology, Linköping University, Sweden \\
  $^2$~Dept. of Computer Science and Technology, University of Cambridge, UK \\
}


\maketitle
\ificcvfinal\thispagestyle{empty}\fi

\begin{abstract}
Single-image high dynamic range (SI-HDR) reconstruction has recently emerged as a problem well-suited for deep learning methods. Each successive technique demonstrates an improvement over existing methods by reporting higher image quality scores. This paper, however, highlights that such improvements in objective metrics do not necessarily translate to visually superior images. The first problem is the use of disparate evaluation conditions in terms of data and metric parameters, calling for a standardized protocol to make it possible to compare between papers. The second problem, which forms the main focus of this paper, is the inherent difficulty in evaluating SI-HDR reconstructions since certain aspects of the reconstruction problem dominate objective differences, thereby introducing a bias. Here, we reproduce a typical evaluation using existing as well as simulated SI-HDR methods to demonstrate how different aspects of the problem affect objective quality metrics. Surprisingly, we found that methods that do not even reconstruct HDR information can compete with state-of-the-art deep learning methods. We show how such results are not representative of the perceived quality and that SI-HDR reconstruction needs better evaluation protocols.
\end{abstract}

\section{Introduction}
Deep learning for high dynamic range (HDR) image reconstruction has gained a great deal of attention over the last few years~\cite{Kalantari2017, Eilertsen2017, Endo2017, Kalantari2019, Chen2021}. With the flurry of recently published papers, it is necessary to have consistent comparisons between them, to ensure progress in a meaningful direction. The de facto standard is to run reference-based objective metrics where reconstruction quality is measured against reference HDR images. 
Most papers use objective metrics like peak signal-to-noise ratio (PSNR), and a fractional increase is deemed sufficient to produce state-of-the-art results.
The objective metrics are usually also accompanied by example image comparisons for showcasing improved reconstruction performance. However, the example images selected by authors are not very representative of overall performance. Due to the stochastic nature of deep learning, different methods will be better on different test images and the objective metrics carry the most weight in demonstrating improvements over previous work.

This paper inspects the commonly used evaluation protocol for single-image HDR (SI-HDR) reconstruction. We start by outlining the SI-HDR reconstruction pipeline and highlight problems with objective evaluations. Next, we discern problems in the high variability in evaluation conditions between papers, where different data, pre-processing, and metric calibration make it impossible to compare results. Furthermore, we point out how differences in the intended use of an LDR-to-HDR method make it difficult to interpret the results. Our primary focus is on the nature of the SI-HDR reconstruction problem itself -- while SI-HDR methods typically claim better reconstruction of under- or over-exposed pixels, the predictions of quality metrics are predominantly affected by the secondary task of inverting the camera response function (CRF). This results in a strong bias towards methods that perform more accurate inversion. 


To summarize our contributions, the outline of the paper is as follows:
Section~\ref{sec:related_work} briefly summarizes the rapidly growing area of deep learning for HDR image reconstruction and points out the lack of structure in evaluations of deep SI-HDR reconstruction methods.
Section~\ref{sec:problem_formulation} outlines the SI-HDR pipeline, with the different sub-problems of SI-HDR reconstruction, and provides a necessary distinction in the intent of single-image LDR-to-HDR methods.
Section~\ref{sec:evaluation_problem} explains three fundamental problems in the evaluation of SI-HDR reconstruction methods.
Finally, Section~\ref{sec:evaluation} compares artificially constructed SI-HDR methods with state-of-the-art methods to highlight how the different problems in Section~\ref{sec:evaluation_problem} affect the outcome of an objective evaluation. We point to how the results most often are not representative of the perceived reconstruction quality.

\section{Related work}\label{sec:related_work}

Early methods combined images of varying exposure to increase the dynamic range \cite{Mann1994, Debevec1997}, but they only work with static scenes. The ability to merge dynamic scenes with misaligned exposures was first provided by optical-flow or patch-matching algorithms~\cite{Kang2003,Sen2012,Tursun2015}. With the advent of deep learning, more challenging scenes with large motions were merged to produce HDR images~\cite{Kalantari2017,Wu2018,Yan2019}.



Concurrently, single-image HDR methods were introduced to display conventional images on HDR displays after applying inverse tone-mapping operators (iTMOs)~\cite{Banterle2006,Meylan2006,Rempel2007,Banterle2008}. Estimating HDR pixel values from a single image is an under-constrained problem due to missing information in under and over-exposed regions. Recently, deep neural networks have demonstrated significant improvements based on a learned high-level understanding of the image content. Following some breakthrough works~\cite{Zhang2017,Endo2017,Eilertsen2017}, a large number of methods have been proposed, each incorporating a different architecture and training strategy~\cite{Marnerides2018,Lee2018,Khan2019,Li2019,Liu2020,Santos2020}. The improved results have resulted in single-image LDR-to-HDR methods being applicable in areas outside HDR display, such as image-based lighting (IBL) and post-processing. As we discuss in \secref{intent}, it becomes important to distinguish between different intents when evaluating methods.

Each previous paper on deep SI-HDR provides a separate evaluation using full-reference objective image quality metrics. Differences in testing data, camera simulation, and metric calibration lead to a great degree of variation between the reported results. Thus, methods considered in several such evaluations show significant variations in the reported quality values, making it impossible to draw general conclusions across different studies.

To the best of our knowledge, there has been only one previous attempt at an independent and standardized evaluation of SI-HDR methods~\cite{perez2021ntire}. This evaluation accompanied the recent HDR challenge in the New Trends in Image Restoration and Enhancement (NTIRE) CVPR workshop on single and multi-image HDR reconstructions with deep-learning methods.
They introduced a new HDR dataset consisting of real as well as synthesized images and objectively evaluated the
reconstructions using PSNR on linear HDR pixels. Additionally, they compared methods on tone-mapped images using the $\mu$-law algorithm. Although the challenge is a step in the right direction towards standardized SI-HDR evaluation, it does not address the problems with SI-HDR evaluation that we focus on in this paper, such as the incorrect use of HDR quality metrics, CRF dominance, and sensitivity to camera simulation.


\section{Problem formulation}\label{sec:problem_formulation}
The forward model for HDR reconstruction problems describes how a camera obtains a low dynamic range (LDR) image, $L$, from an HDR ground truth image, $H$,
%
\begin{equation}
    L(x) = q\left( \min\{1, g(e\,H(x) + \eta(H(x)))\} \right),
    \label{eq:cs}
\end{equation}
where $x$ denotes pixel index, $e$ is a simulated exposure time, $\eta$ is signal-dependent camera noise \cite{Hanji2020}, $q$ performs quantization to the desired bit depth, and $g$ is a non-linear tone-mapping or camera response function (CRF).
Transforming a low dynamic range image to high dynamic range (LDR-to-HDR) aims at approximating the inverse of \eqref{cs}. This is a multi-faceted problem that can be divided into the following sub-problems:
\begin{itemize}
    \item[R-I] \textbf{Inversion of the CRF --} Estimating the inverse $g^{-1}$ is a challenging problem and is impossible without the presence of enough contextual information in an image. At the same time, one can argue that this problem is not the most important in SI-HDR reconstruction for at least two reasons. Firstly, a perfect linearization/CRF inversion is not necessary for most applications. This is because some applications are not sensitive to linearization, and for others, some non-linearity will be applied in the end anyway. Secondly, for many applications, linear data is readily available (e.g., RAW images directly from the camera sensor).
    
    
    \item[R-II] \textbf{Bit-depth expansion/de-quantization --} Recovering higher bit-depth is typically unnecessary for modern cameras since they are equipped with analog-to-digital converters with sufficient bit-depth. However, when a CRF with shallow slopes needs to be inverted, the process may introduce visible quantization artifacts.
    
    
    \item[R-III] \textbf{Reconstructing of under-exposed pixels --} These pixels are mostly affected by noise, $\eta$ in \eqref{cs}, and in most cases this problem is similar to image denoising. The details in dark regions can also be lost due to quantization, for example if the pixels are largely under-exposed and/or aggressively compressed by a CRF.
    
    
    \item[R-IV] \textbf{Reconstruction of over-exposed pixels --} The clipping of information due to sensor saturation, modeled by $\min\{\ldots\}$ in \eqref{cs}, is most often directly associated with the SI-HDR reconstruction problem. While the problem shares some similarities with the recovery of under-exposed pixels, it also differs significantly in many aspects. Perhaps the most relevant difference is the fact that over-exposed pixels often carry most of the dynamic range. This results in a long-tailed distribution of luminances because specular highlights and light sources are much brighter than other parts of a scene. For some applications, such as IBL, this information is critical. The problem is also significantly different in terms of solution and is more closely related to inpainting. 
    
\end{itemize}

\subsection{HDR reconstruction intent}\label{sec:intent}
\begin{figure}
    \centering
    \includegraphics[width=\linewidth]{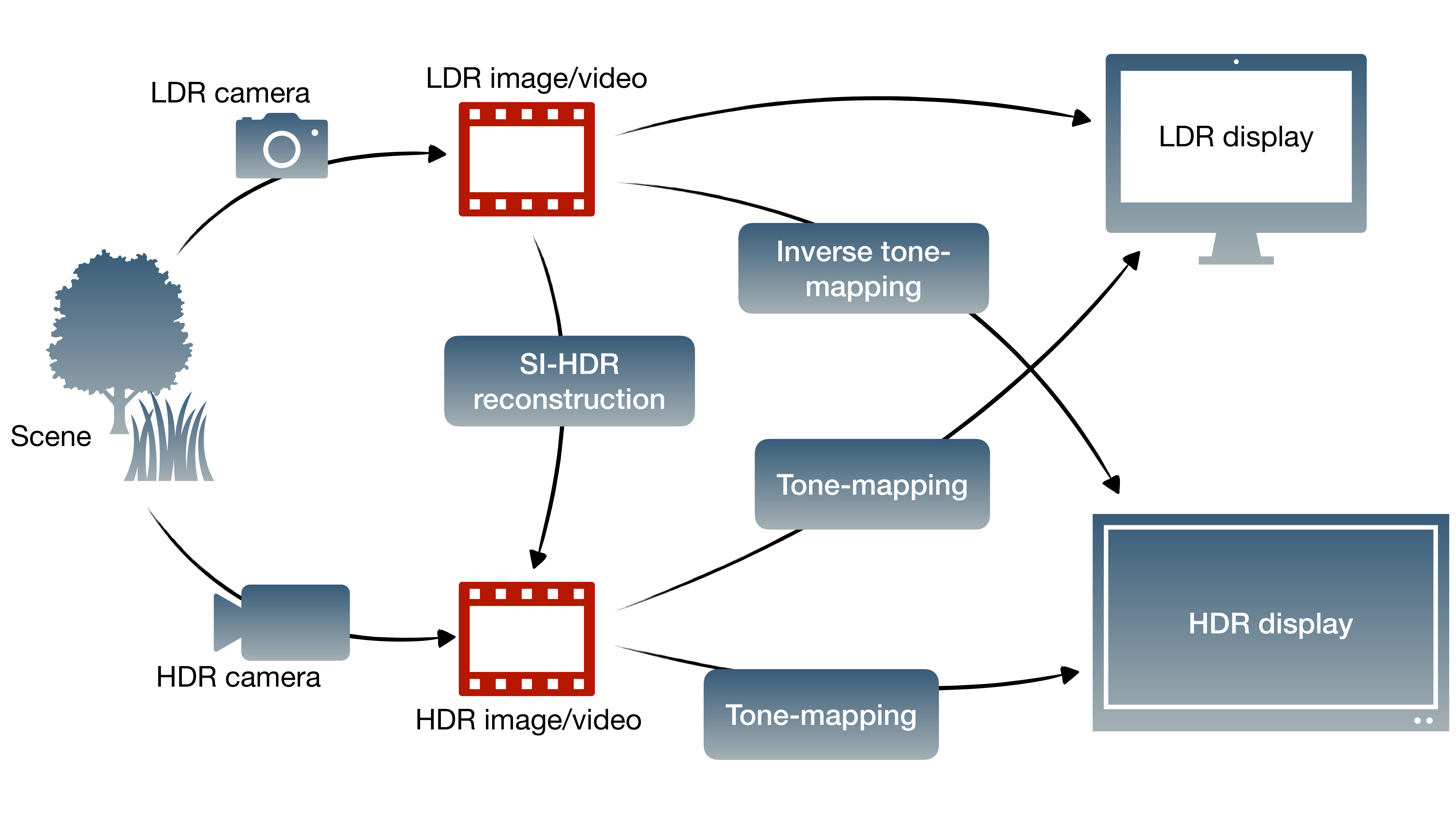}
    \caption{Illustration of how the intent of SI-HDR reconstruction and inverse tone-mapping (iTM) differs. The figure shows LDR/HDR display as the direct application, but the HDR images could also be used for other purposes (IBL, post-processing, etc.). While iTM directly maps LDR content for HDR display, a similar goal could be achieved by SI-HDR followed by tone-mapping.}
    \label{fig:illustration}
\end{figure}

The SI-HDR problem is typically understood as the inverse problem of \eqref{cs}. A closely related problem termed \emph{inverse tone-mapping} (iTM) aims at enhancing LDR image $L$ but without the goal of matching the original HDR image $H$. 
The difference between SI-HDR and iTM is illustrated in Figure~\ref{fig:illustration}, and can be defined as follows:

\pgraph{Inverse tone-mapping (iTM)} Methods that use an inverse tone-mapping operator (iTMO) for maximizing subjective quality on an HDR display. In this scenario, color gamut, exposure, and local contrast should be correctly tuned for an improved viewing experience. Recovering information in highlights is not essential since a rich viewing experience is possible even if some information is clipped to the peak luminance of the display. This category includes the earlier heuristic methods~\cite{Banterle2006,Meylan2006,Rempel2007,Banterle2008}, as well as more recent learning-based models~\cite{Kim2019,Kim2020}.

\pgraph{Single-image HDR (SI-HDR) reconstruction} Methods that aim at recovering the underlying physical light quantities of the captured scene. In this scenario, one of the most important problems is to recover lost information in the saturated areas of the image (R-IV). 

It should also be noted that an iTMO can be composed by taking the result of an SI-HDR method and tone-mapping it for the HDR display, thus potentially disregarding some of the reconstructed information.
In previous work, we recognize a certain confusion between the two intents, where some methods are not clear on which objective they aim for, and where iTM is used to describe methods that aim for HDR reconstruction. 

\section{The evaluation problem}\label{sec:evaluation_problem}
The evaluation of iTM is similar to that of conventional tone-mapping, which aims at displaying HDR images on LDR monitors. That is, it is difficult to formulate a numerical evaluation measure that reflects subjective preferences. 
The evaluation of SI-HDR reconstruction, on the other hand, is closer to multi-exposure HDR reconstruction where the underlying luminance values of the scene are sought.
A major problem with recent SI-HDR papers is confusion in intent, causing methods to be presented as iTMOs but evaluated by comparing them to reference HDR images.
This, by itself, imposes difficulties in properly assessing a method. However, even if a method is clearly performing SI-HDR reconstruction, it is still problematic to numerically evaluate its performance, for a number of reasons. We focus on three problems that obscure proper comparisons between methods: \emph{CRF dominance}, \emph{CRF bias}, and \emph{proper use of HDR metrics}.




\subsection{CRF dominance}\label{sec:crfd}
For a deep-learning SI-HDR reconstruction, a neural network $f$ is trained to invert \eqref{cs}, providing the estimate $\hat{H} = f(L)$. A natural choice for the evaluation of different methods is to compare the reconstruction quality in terms of the difference between a reconstructed HDR image and the corresponding ground truth, $d(\hat{H},H)$. This difference is typically an HDR quality metric (discussed in \secref{hdr-metrics}).
The main problem with this approach is that the numerical difference between $\hat{H}$ and $H$ is usually dominated by inaccuracies in inverting $g$ (R-I), which affects all pixels in the image. Failure to properly recover clipped pixel values due to under- or over-exposure (R-III and R-IV) has a relatively small impact, especially for the moderate amount of saturation found in most images. Thus, a method that accurately estimates $g^{-1}$ is likely to perform significantly better in an objective evaluation compared to one that provides a good reconstruction of over and under-exposed pixels.
In the experiment in \secref{evaluation} we will demonstrate this problem.



%
%

\subsection{CRF bias}
The testing datasets of most SI-HDR papers contain images with a similar formulation of CRFs as the training data.
However, there is no guarantee that contemporary methods were trained with similar CRFs, so these can potentially be less suitable for inverting the specific CRFs used in the evaluation. Thus, it is easy to create a bias towards the proposed method and report improved quantitative results. Due to CRF inversion being a dominant feature of the inversion problems (Section~\ref{sec:crfd}), this type of bias is expected to have a significant influence on the results of an evaluation.


As an extreme case, consider using a single CRF for training a deep learning method for SI-HDR reconstruction and using the same CRF for the testing data. Now, the method only needs to learn a single linearization function, $g^{-1}()$, which is a relatively simple problem.
When compared against other methods that have been trained on a variety of CRFs, the proposed method will have a strong advantage.
In most previous works, the problem is not as extreme because LDR data is usually simulated by a range of different CRFs. However, since camera simulation is usually formulated differently in different papers, there could still be a bias due to a closer similarity between training and testing CRFs. 

\subsection{HDR metrics}
\label{sec:hdr-metrics}

\begin{figure}
    \centering
    \includegraphics[width=\columnwidth]{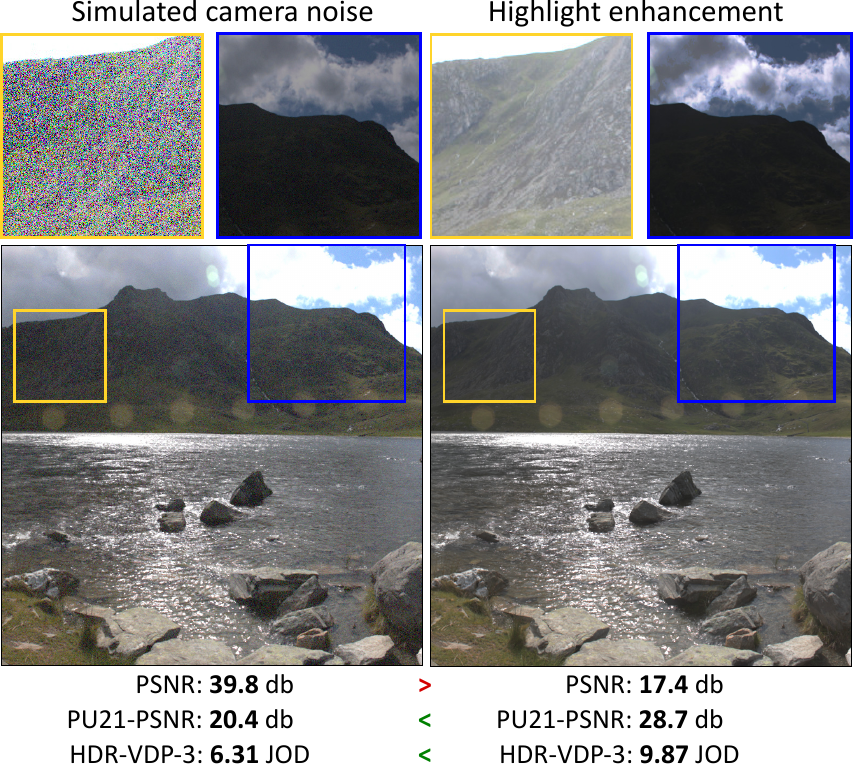}
    \caption{Quality predictions of regular (PSNR) and HDR image quality metrics (PU21-PSNR and HDR-VDP-3) on an HDR image in linear color space. The image contains either camera noise (left) or a highlight enhancement (right). PSNR incorrectly predicts that image quality is higher for the image affected by noise. This is because PSNR computed on linear color values is very sensitive to changes in the bright parts of an image and is much less sensitive to changes in dark parts, which are most affected by camera noise. The HDR image was gamma-encoded ($\gamma=2.2$) for presentation. Different exposures are shown in the insets.}
    \label{fig:hdr-metric-pred}
\end{figure}

The final consideration is the proper use of HDR quality metrics. The images generated by SI-HDR methods are typically represented in linear RGB color spaces, which lack the perceptual uniformity of display-encoded color spaces, such as sRGB. For this reason, standard quality metrics, such as PSNR or SSIM, cannot be used to assess the quality of reconstructed HDR images. The problem is illustrated in \figref{hdr-metric-pred}, in which an image with a high amount of camera noise has a much higher PSNR than the same image with enhanced highlights (contrast stretched in bright areas). Such a prediction contradicts what we see in the image in which the noise is much more apparent than highlight enhancement. This is because the pixel differences in linear color spaces emphasize the differences in bright areas and de-emphasize the differences in dark areas. To provide a meaningful prediction of image quality, we can either use a dedicated HDR quality metric, such as HDR-VDP \cite{Mantiuk2011}, or transform an image into a uniform color space using perceptually uniform transform, such as PU21 \cite{Mantiuk2021}. 

Another common problem is the improper use of HDR metrics. Unlike standard metrics, HDR metrics require images to be represented in absolute colorimetric values, which correspond to the light emitted from a display. This is because such metrics account for the fact that the visual system is less sensitive to dark (low luminance) colors. To compute meaningful predictions for an HDR image in a linear color space, its exposure needs to be adjusted by multiplying pixel values by a constant. One common strategy is to map the peak value of an image to 1000 (for 1000\cdms display), and another is to map diffuse white color to 200 (200\cdms is typical luminance of the white for a monitor with a comfortable level of brightness). For reproducibility, the strategy for performing such a mapping and the metric parameters used (display size, resolution, viewing distance) should be reported in each paper. For example, the results for \figref{hdr-metric-pred} were computed assuming a 24-inch display with a resolution of 1920$\times$1200 pixels, at a viewing distance of 0.5\,m and the peak luminance of the image mapped to 400\cdms.


\begin{figure}
    \centering
    \includegraphics[width=0.78\linewidth]{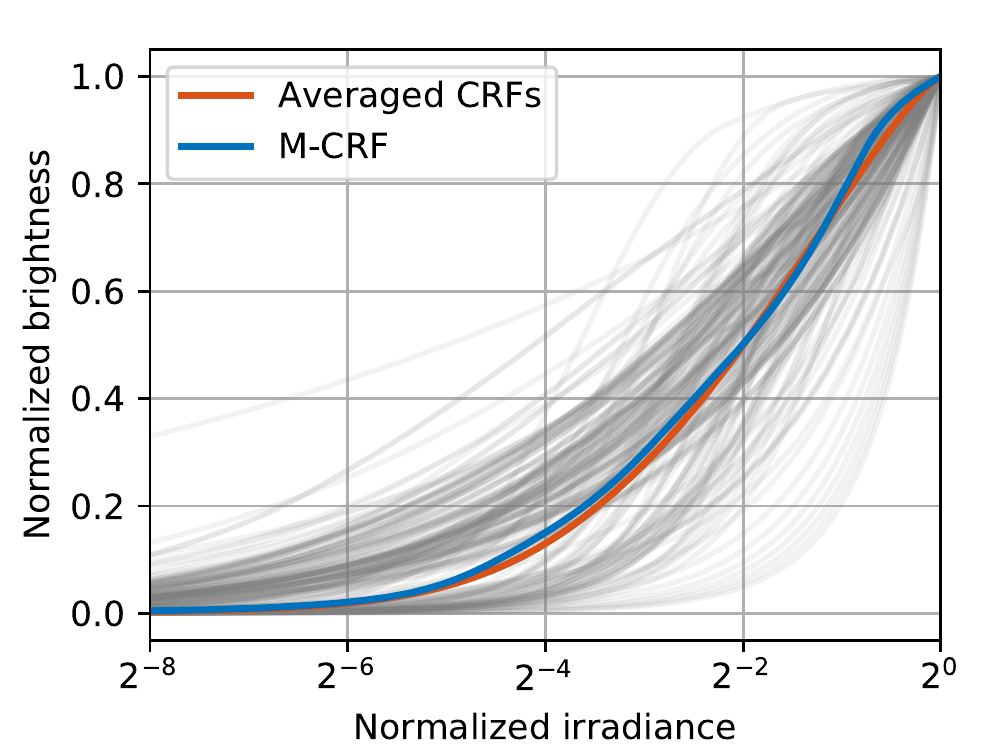}
    \caption{The selected response function, M-CRF, from the database in~\cite{Grossberg2003}, closely similar to the mean of CRFs. The grey curves are all the 201 CRFs of the database.}
    \label{fig:crfs}
\end{figure}

\begin{figure*}
    \centering
    \begin{subfigure}[b]{\textwidth}
        \centering
        \includegraphics[width=\linewidth]{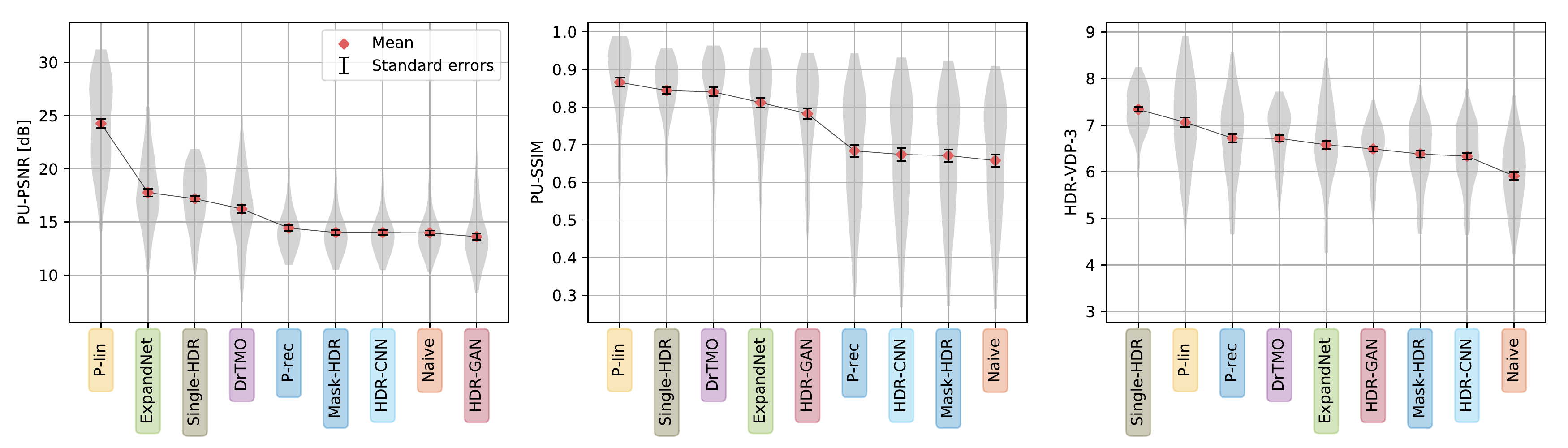}
        \caption{Camera simulation: M-CRF, EV-5}
    \end{subfigure}
    \begin{subfigure}[b]{\textwidth}
        \centering
        \includegraphics[width=\linewidth]{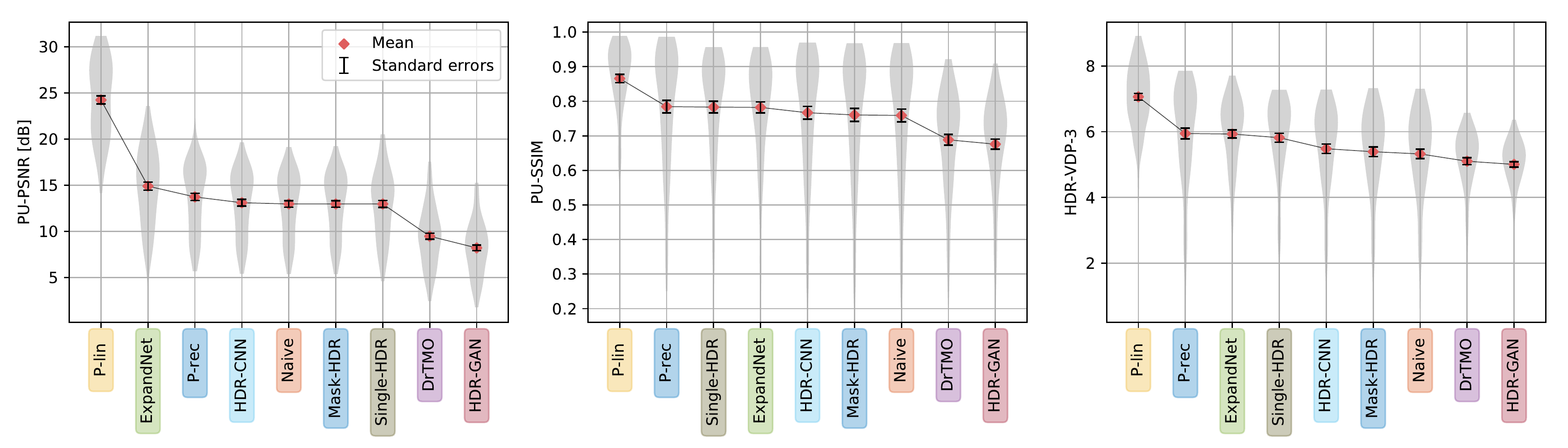}
        \caption{Camera simulation: CLAHE, EV-5}
    \end{subfigure}
    
    \caption{The distribution of metric values over the 96 tested scenes, where methods have been color-coded and sorted by mean value to facilitate comparing differences in ranking. (a) uses camera simulation with M-CRF, while (b) is with CLAHE, and both have been simulated with EV-5. Left, middle, and right show results with PU-PSNR, PU-SSIM, and HDR-VDP-3, respectively. The methods are: DrTMO~\cite{Endo2017}, HDR-CNN~\cite{Eilertsen2017}, ExpandNet~\cite{Marnerides2018}, HDR-GAN~\cite{Lee2018}, Single-HDR~\cite{Liu2020}, and Mask-HDR~\cite{Santos2020}, as well as 3 reference methods (see \secref{experimental_setup} for details). Error bars denote standard errors.}
    \label{fig:eval}
\end{figure*}

\section{Evaluation}\label{sec:evaluation}
For demonstrating the aforementioned problems with CRF dominance and CRF bias, we set up an evaluation that is formulated to highlight the influence of CRF inversion. We do this both by testing individually using different CRF formulations and by creating artificial SI-HDR methods that can do either perfect CRF inversion or perfect reconstruction of saturated pixels.

\subsection{Experimental setup}\label{sec:experimental_setup}

\pgraph{Camera simulation}
We simulate cameras using two different CRFs. The first one, called M-CRF, is a static CRF selected from a popular database~\cite{Grossberg2003}. It was selected as the CRF closest to the mean CRF over the set of 201 CRFs (see \figref{crfs}). This CRF is representative of what was used when training most SI-HDR methods.
The second one, called CLAHE, is an adaptive CRF that implements Contrast Limited Adaptive Histogram Equalization~\cite{Pizer1987} and produces a different CRF for each image. We compute image histograms and create a mapping on logarithmic luminance values to ensure better perceptual uniformity.
The method is chosen to be representative of the image processing utilized in modern smartphone cameras.
For proper sensor simulation, we also include noise simulation according to~\cite{Hanji2020}, which utilizes parameters measured from a Canon EOS-1Ds with exposure time 1/30 sec and ISO~800.
For each HDR scene, we simulate images at two different exposure value (EV) settings, EV-5 and EV-10, which are tuned individually for each image to clip 5\% and 10\% of the brightest pixels, respectively. We use these to test the methods' abilities to perform reconstruction in scenarios of varying difficulty, where a higher degree of clipped pixels will make SI-HDR reconstruction more challenging.

\pgraph{HDR metrics}
For measuring objective reference-based quality, we use three common metrics. To use LDR metrics PSNR and SSIM, we transform RGB images to perceptually uniform units with PU21 (discussed in \secref{hdr-metrics}).
Additionally, we use the latest HDR-VDP~\cite{Mantiuk2011,Narwaria2015} release, HDR-VDP-3, and report the Q\_JOD-score quality correlate defined in the range $[0,10]$. For each comparison, we scale the clipping points for EV-5 and EV-10 of each image to 500\cdms, i.e., the 95th and 90th percentiles are anchored to 500\cdms, for EV-5 and EV-10, respectively. For HDR-VDP-3 we specify a 24-inch display with a resolution of 1920$\times$1200 pixels, at a viewing distance of $0.5$ meters.

\pgraph{Methods}
For comparisons, we focus on 6 of the most popular and recent SI-HDR reconstruction methods in the literature, each used with pre-trained weights provided by respective authors. 
The methods are: DrTMO~\cite{Endo2017}, HDR-CNN~\cite{Eilertsen2017}, ExpandNet~\cite{Marnerides2018}, HDR-GAN~\cite{Lee2018}, Single-HDR~\cite{Liu2020}, and Mask-HDR~\cite{Santos2020}.

To test the influence of CRF specification, we also create two artificial SI-HDR methods. The method \emph{P-lin} performs perfect linearization (problem R-I), but no reconstruction of saturated pixel information,
\begin{equation}
\hat{H}_\textrm{P-lin}(x) = q\left( \min\{1, eH(x) + n(H(x))\} \right).
\end{equation}
That is, P-lin is a simulated LDR image with no CRF applied. Although noise and quantization are included, comparing this image to the ground truth $H$ will mainly measure the differences caused by missing information due to over-exposure (problem R-IV).
The second model, \emph{P-rec} performs perfect reconstruction of the saturated pixels but uses an imperfect static inverse CRF $L_\textrm{lin}=L^2$,
\begin{equation}
\hat{H}_\textrm{P-rec}(x) = \alpha(x) H(x) + (1-\alpha(x))L(x)^2,
\end{equation}
where $\alpha$ is a mask that extracts saturated regions in the image, $\alpha = \max(0, L-0.9)/0.1$. This means that the information for over-exposed pixels will be taken from the ground truth image $H$.
Finally, we also include a naive model that performs no reconstruction and imperfect linearization, $\hat{H}_\textrm{naive} = L^2$. We include this to highlight how doing nothing in some circumstances can outperform state-of-the-art SI-HDR reconstruction methods. 

\begin{figure}
    \centering
    \includegraphics[width=\linewidth]{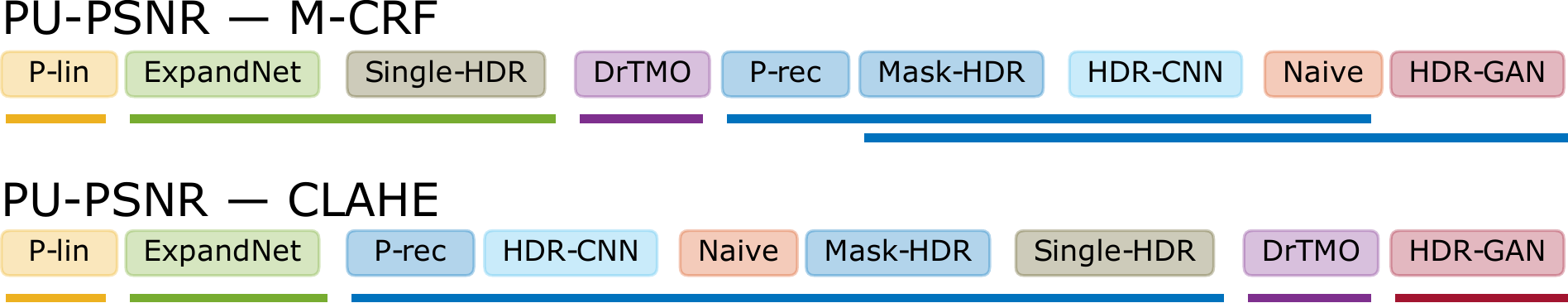}
    \caption{The rankings provided by PU-PSNR with M-CRF and CLAHE (the best method on the left). The lines connect methods where differences cannot be deemed statistically significant in a t-test, with a p-value threshold of $0.05$.}
    \label{fig:rankings}
\end{figure}

\pgraph{Data}
We use 96 images from the test set in~\cite{Eilertsen2017}, which is a combination of online HDR resources as well as captured HDR scenes. Although there is a potential risk that some of the images have been used in training by some of the compared methods, we regard this aspect as having a minor influence on the outcome of the experiments -- the overlap between this set and the methods' training data should be small, and there are significant differences in camera simulation and cropping. All images are 1024$\times$768 pixels.

\subsection{Results}
The results separated between each metric and CRF are presented in Figure~\ref{fig:eval}.
We sort the methods according to the mean performance to highlight how the ranking between methods changes for different CRFs.
Since the performance varies significantly between scenes, the distributions of metrics are often wide, but standard errors show that the 96 images provide a good estimate of the mean. To highlight which of the methods are possible to separate in the rankings, \figref{rankings} shows the rankings for PU-PSNR. The lines connect methods where a t-test cannot reject the null hypothesis that methods come from the same distribution, at the 5\% significance level.
The differences in performances between some methods are not statistically different, but overall there is a clear difference between low and high-performing methods.
For results using all combinations of camera simulation and evaluation metrics, we refer to the supplementary material.

\begin{figure}
    \centering
    \includegraphics[width=0.9\linewidth]{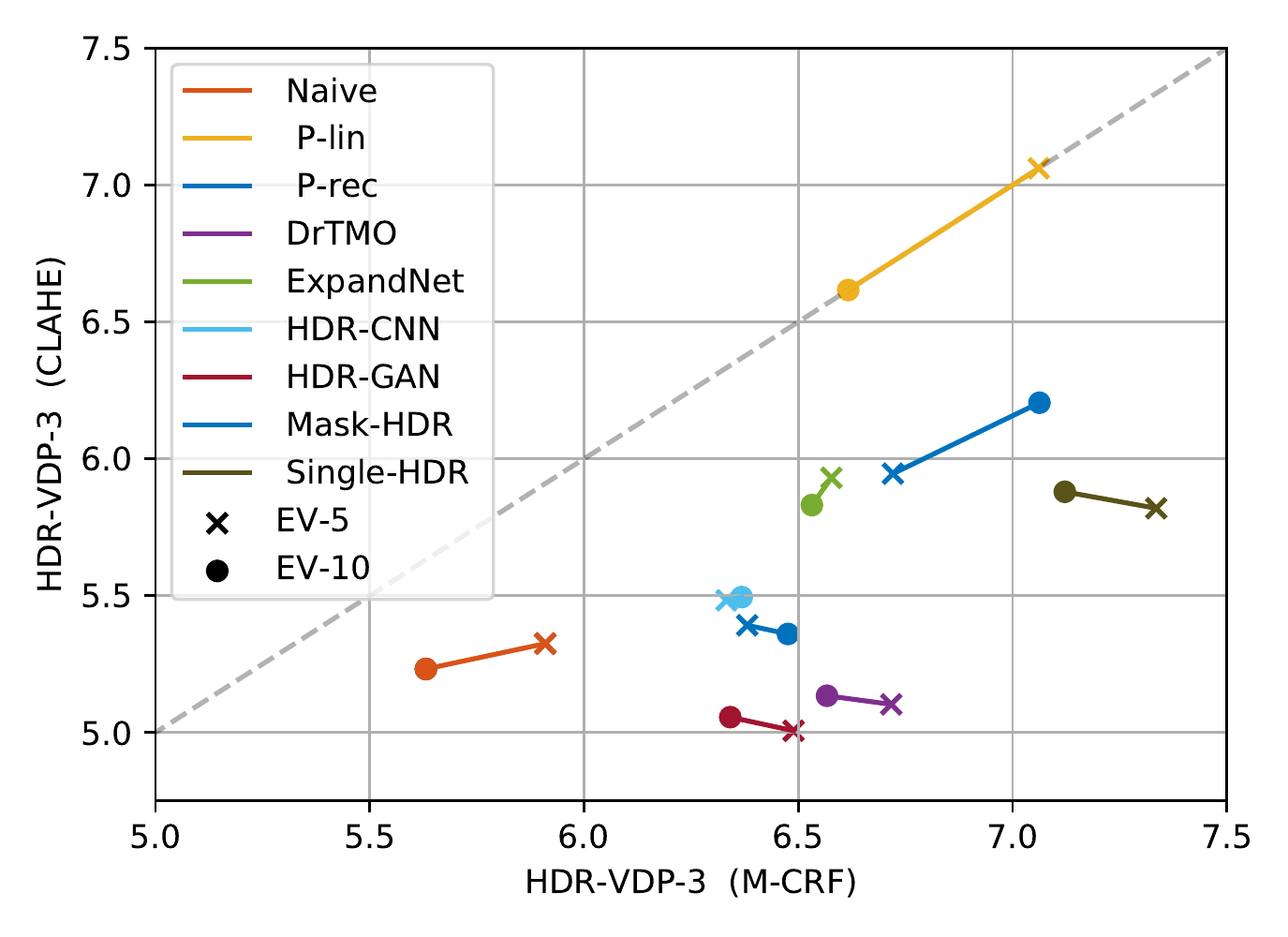}
    \caption{Differences in HDR-VDP-3 when using M-CRF and CLAHE. The two points for each method are with EV-5 and EV-10, demonstrating how several methods do not show an expected reduction in quality with increased camera exposure (more challenging reconstruction problem).}
    \label{fig:crf_ev}
\end{figure}

\begin{figure*}
    \centering
     \includegraphics[width=0.996\linewidth]{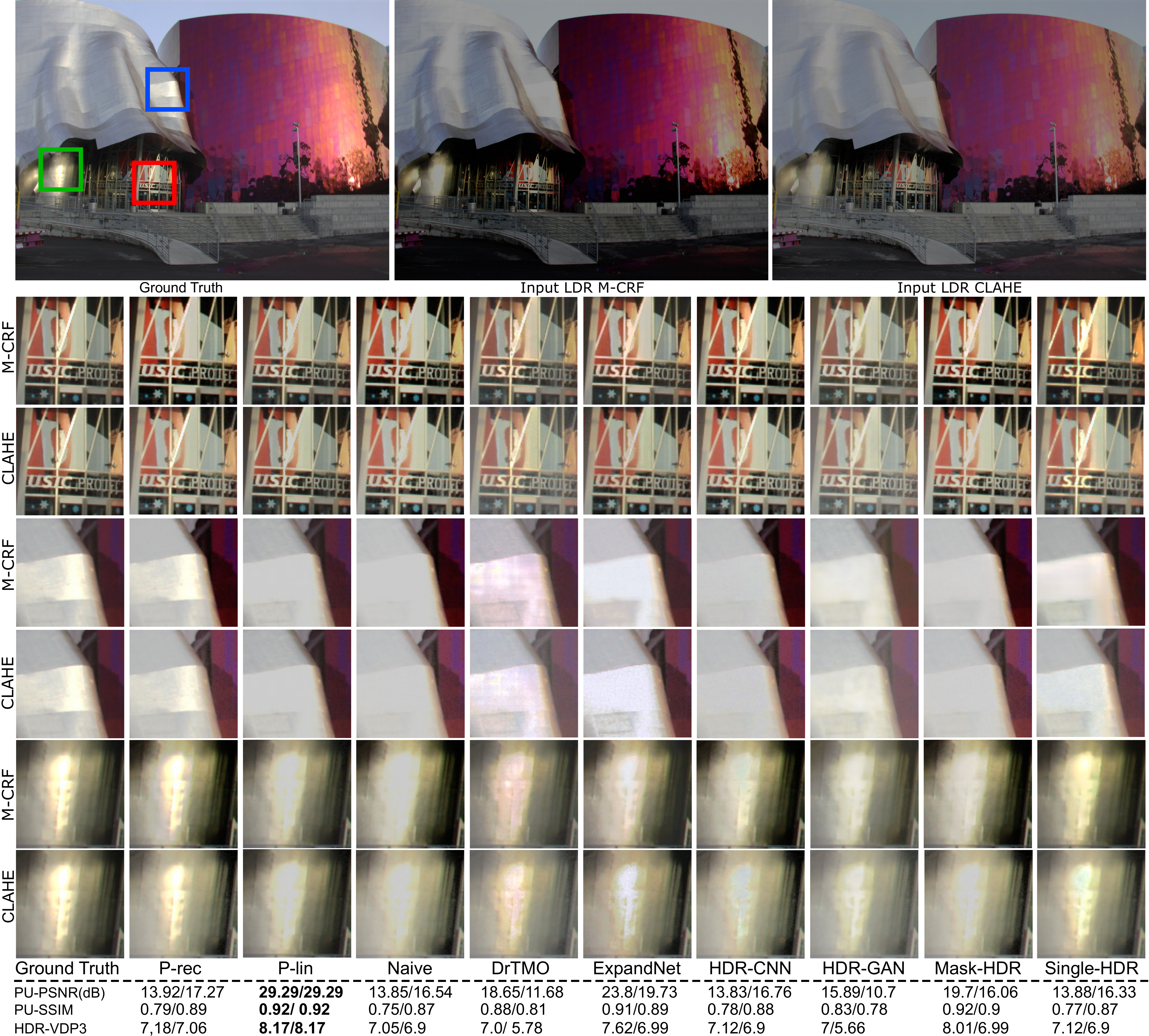}
    \caption{Selected scene areas for different reconstructions, with input LDR images simulated using M-CRF and CLAHE. The metrics at the bottom show the performance with M-CRF/CLAHE. The exposure time was set such that $5\%$ of pixels are saturated (EV-5). Ground truth and reconstructed HDR images have been gamma-encoded for display.}
    \label{fig:comparison}
\end{figure*}

\begin{figure*}
    \centering
    \includegraphics[width=\linewidth]{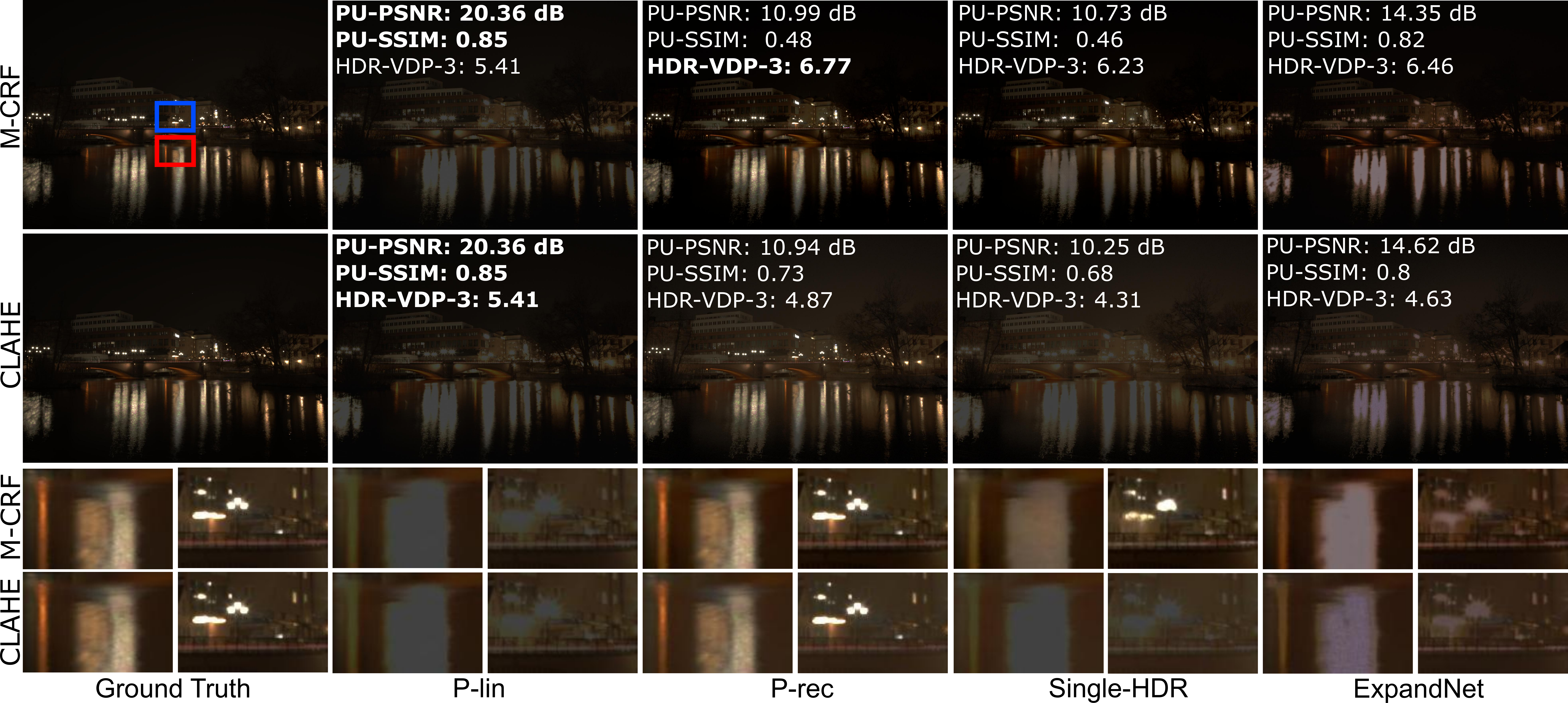}
    \caption{Comparison of a selection of reconstruction methods, on input LDR images simulated with M-CRF and CLAHE. Camera simulations have been performed using EV-5. Images have been gamma-encoded for display, and the exposure has been reduced by -2 EV to facilitate comparisons in bright image areas.}
    \label{fig:fullsizeComp}
\end{figure*}

The most notable pattern is how the P-lin method outperforms all other methods with a large margin. This is consistent across different combinations of camera simulation, EVs, and evaluation metrics, except for M-CRF with HDR-VDP-3. Since P-lin only inverts the CRF and does not reconstruct HDR information, the results indicate that the dominant feature of the reconstruction problem is the linearization of pixel values.
Also, this means that if a model is trained and tested in a way that gives it an advantage in terms of CRF inversion quality, it will be easy to demonstrate improvements over previous work without having to perform successful reconstruction of saturated pixels.

The ranking between the compared methods is sensitive to the formulation of camera simulation. For example, DrTMO and the naive model have significantly different performances.
DrTMO is top 3-4 for M-CRF, while the naive model shows the worst performance. On the other hand, with CLAHE, the naive model outperforms DrTMO and is comparable to the bulk of state-of-the-art methods (see \figref{rankings}). As the naive model does not perform HDR reconstruction or estimation of the inverse CRF, this demonstrates how easy it is to cheat SI-HDR evaluations with different CRFs.

To better demonstrate the disagreement between evaluations with different camera simulations, \figref{crf_ev} plots the mean HDR-VDP-3 score for M-CRF against the score for CLAHE. This is done both for camera simulations at exposure EV-5 and EV-10, denoted by the different points. For results to be consistent between camera simulations, we expect the evaluation done with either CRF to produce very similar results, but this is not the case. It is apparent how widely the results differ depending on the CRF formulation. Also, comparing the different exposures, they do not have the expected impact on the results.
Since the longer exposure (EV-10) generates a more challenging reconstruction problem with more missing information than the shorter one (EV-5), the quality should decrease. Such a decrease is seen for the P-lin results that are not affected by the CRF. However, on the contrary, for most of the other methods the quality increases for one or both camera simulations.

Connecting the numerical results to the visual quality of reconstructions, \figref{comparison} shows an example scene with reconstructions in selected areas for all methods, while \figref{fullsizeComp} shows an example for a selection of methods in full size. Compared to the ground truth image, it is evident how the P-rec method has the best HDR quality, while P-lin has no visual improvements, which is in stark contrast to objective metrics. Also, the perceived differences between P-lin and the naive method are marginal, despite the differences in measured quality. In \figref{fullsizeComp}, Single-HDR works well for the M-CRF case, while not with CLAHE, but this is hard to decipher from the numbers. ExpandNet, which together with Single-HDR is one of the best-performing methods (except for P-lin), shows high PSNR values but recovers little high-intensity information. The most likely explanation, in this example, is that the method does better CRF inversion compared to other methods.

\section{Conclusion}\label{sec:conclusion}
In this paper, we have focused on common problems with the objective evaluation of SI-HDR methods. We have included a specification of the SI-HDR reconstruction pipeline, pinpointed sub-problems that pose difficulties for evaluation, and highlighted the differences between intents of LDR-to-HDR reconstructions. Our experimental results unveiled a disagreement between the numerical results and the visual quality of reconstructed images. In terms of the perceived quality, the P-rec method, which accurately recovered HDR information in over-exposed image areas, provided the most successful HDR reconstruction. However, object metrics did not reflect this. Instead, they favor the P-lin method, which does not perform HDR reconstruction but outperformed all other methods due to CRF dominance. Moreover, we demonstrated how results are highly sensitive to camera simulation and metric calibration, making it difficult to draw conclusions from such an evaluation without a standardized evaluation protocol. The sensitivity to camera simulation and metric calibration also makes it easy to cheat by formulating the evaluation in a way that favors a particular model.

To make progress in deep LDR-to-HDR, it is crucial to align evaluations for meaningful comparisons between different methods and between different evaluations. 
One strategy for isolating the quality of highlight recovery is to only evaluate differences in saturated areas. However, this is not likely to overcome the problem, as incorrect linearization will also affect bright pixels, i.e., there will still be a significant difference between methods with good and bad linearization, even if these are equally good at reconstructing the missing information (for an example, we refer to the supplementary material). To fully overcome the problem, CRF inversion needs to be more clearly separated. We emphasize the importance of focusing on this in future work.


\pgraph{Acknowledgments}
{\small This project was partially funded by Wallenberg Autonomous Systems and Software Program (WASP), the strategic research environment ELLIIT, and the Knut and Alice Wallenberg Foundation (KAW). It has also received funding from the European Research Council (ERC) under the European Union's Horizon 2020 research and innovation programme (grant agreement N$^\circ$ 725253--EyeCode).}

{\small
\bibliographystyle{ieee_fullname}
\bibliography{egbib}
}

\newpage
\section*{\Large Supplementary material}
\appendix

\section{Dataset and camera simulation}
\begin{figure*}
    \centering
    \includegraphics[width=\linewidth,trim=0 7cm 0 0, clip]{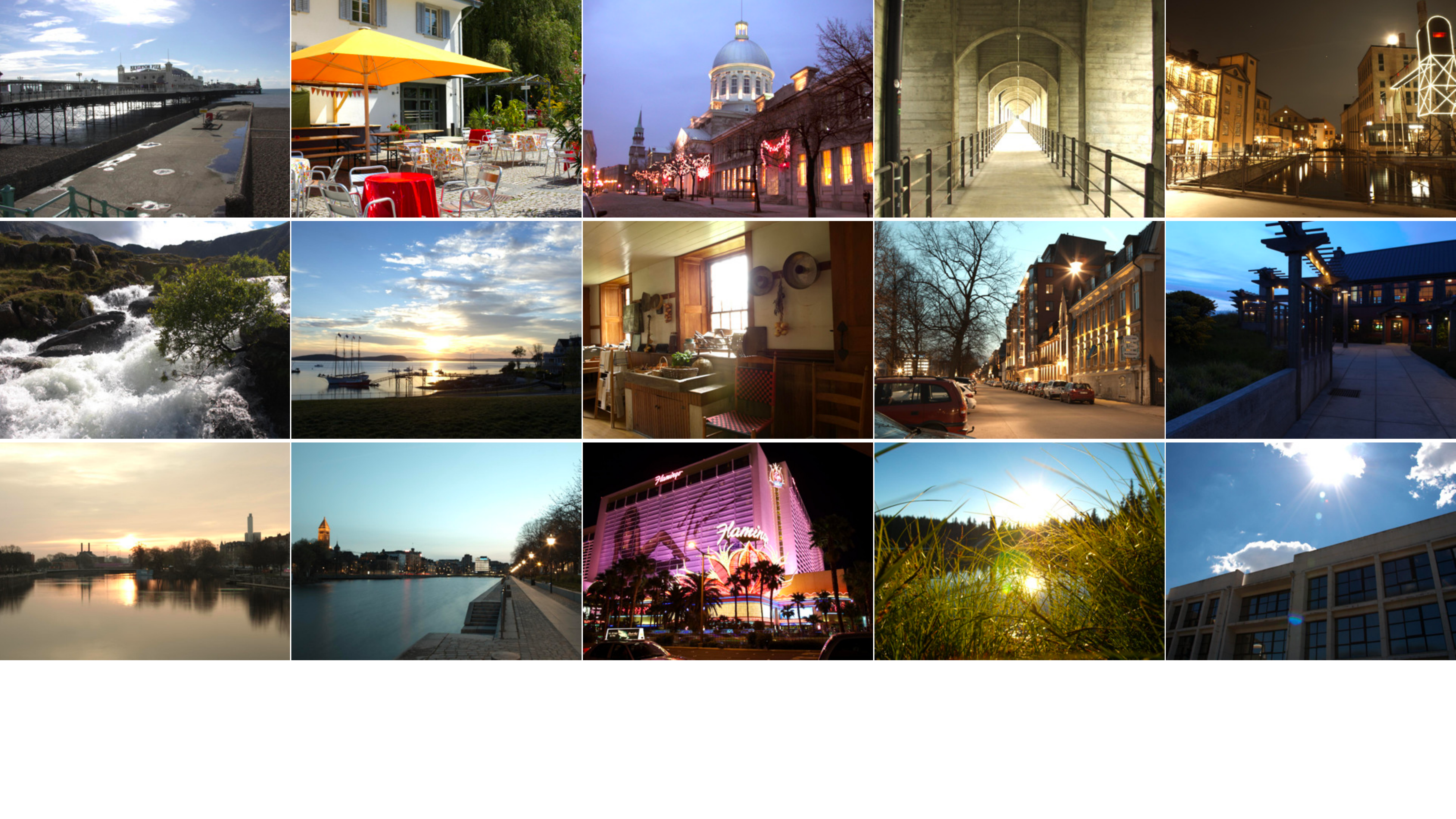}
    \caption{15 example HDR scenes from  the 96 scenes used in evaluation. Images have been gamma-encoded for display.}
    \label{fig:dataset}
\end{figure*}

\begin{figure*}
    \centering
    \begin{subfigure}[b]{0.49\textwidth}
        \centering
        \includegraphics[width=\linewidth,trim=17.4cm 0 0 0, clip]{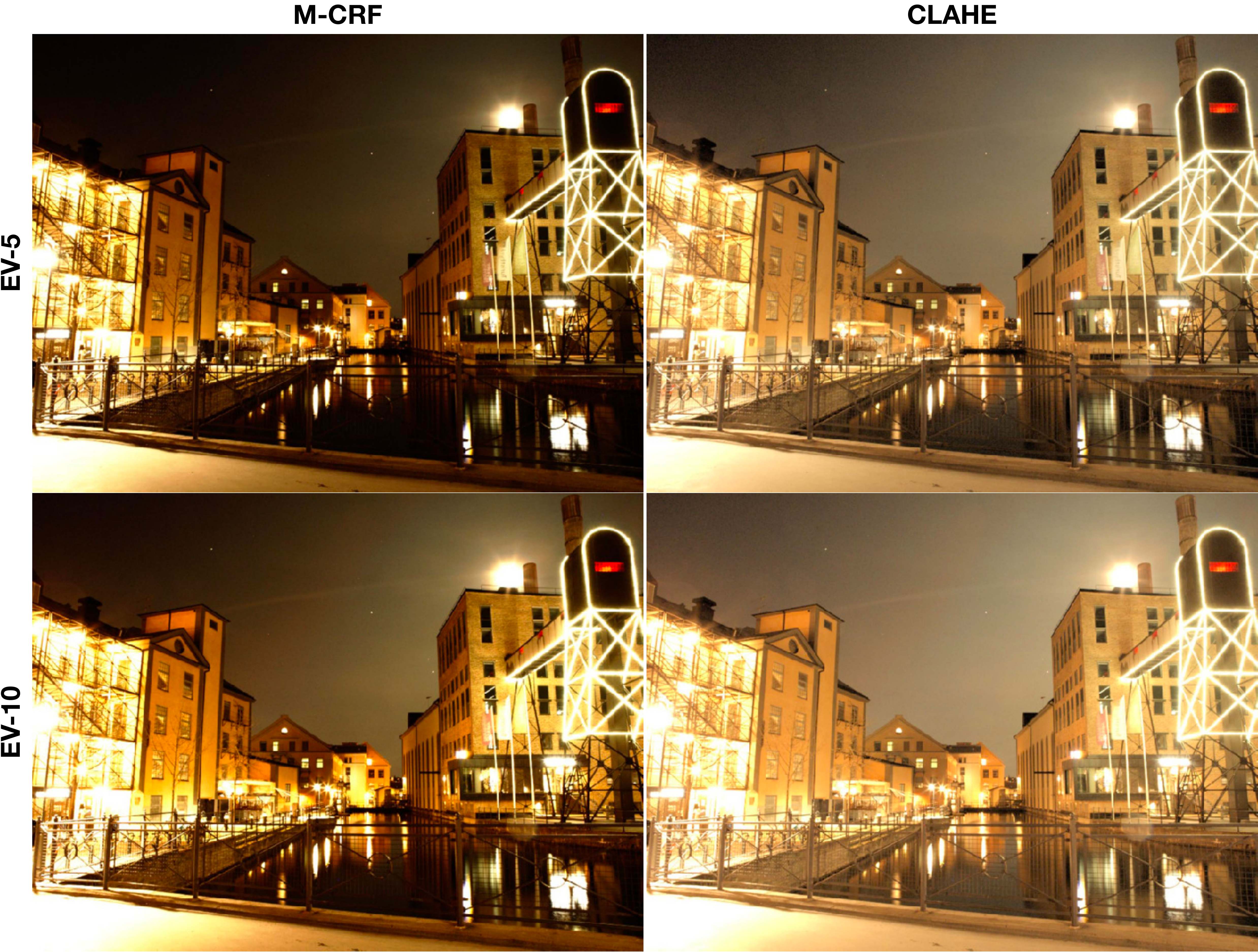}
        \caption{Example 1}
    \end{subfigure}
    \hspace{0.1cm}
    \begin{subfigure}[b]{0.49\textwidth}
        \centering
        \includegraphics[width=\linewidth,trim=17.4cm 0 0 0, clip]{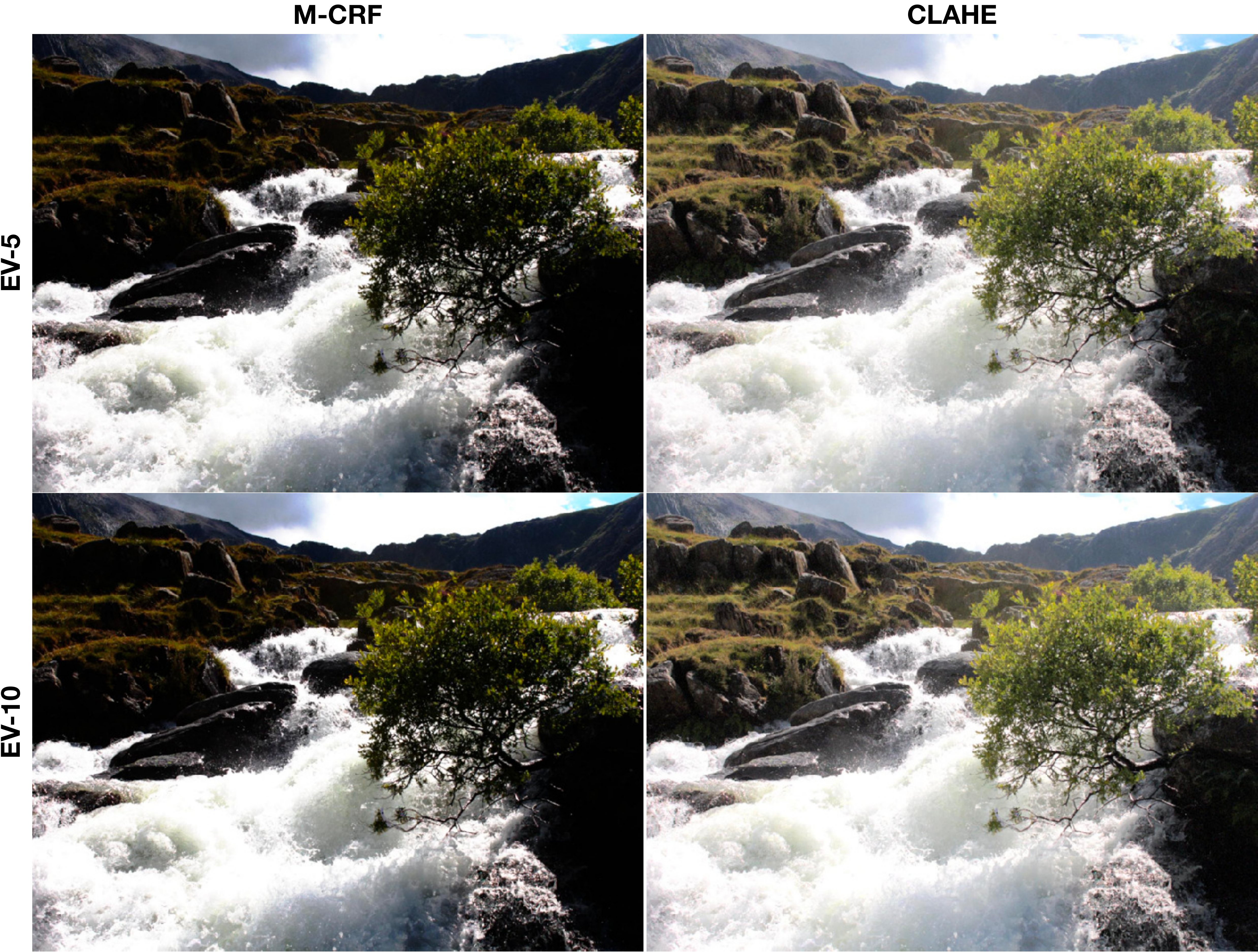}
        \caption{Example 2}
    \end{subfigure}
    \caption{Examples of camera simulation, showcasing the results of different CRFs and exposures used for evaluation.}
    \label{fig:cam_sim}
\end{figure*}

\figref{dataset} shows examples of HDR scenes used in the experiments. \figref{cam_sim} demonstrates the impact of the different camera simulations that were used.
    
\section{Complementing results}
\begin{itemize}
    \item \figref{eval} complements Figure 3 in the main paper, with the same results but for EV-10 instead of EV-5.
    \item \figref{rankings5} and \figref{rankings10} complement Figure 4 in the main paper, with statistical testing of rankings for all metrics and camera simulations.
    \item \figref{crf_ev} complements Figure 5 in the main paper, with plots for all three different metrics.
    \item \figref{comparison2} and \figref{comparison3} complement Figure 6 in the main paper, with similar examples for other scenes.
\end{itemize}

\section{Supplementary results}
\figref{eval_sat} shows the same evaluations as in Figure 3 in the main paper, but evaluated only on saturated pixels. That is, if the metric is $d(\hat{H},H)$, these results have been computed using $d(\alpha\hat{H},\alpha H)$, where $\alpha = \max(0, L-0.9)/0.1$ masks out only the saturated pixels. The rankings with testing of significant differences are given in \figref{rankings_sat}.

The comparison of only saturated regions can clearly separate P-rec as the best model. This is expected, since this has been composed using the ground truth information in the saturated pixels. It is also possible to see how the naive model is inferior, since this does not contain any information in the saturated pixel areas. However, for all other methods it is difficult to clearly separate between methods. Many methods also show differences in ranking depending on camera simulation.

P-lin does not contain any information in saturated regions, similar as with the naive model. Still, P-lin shows slightly higher mean. This is likely due to the blending performed by $\alpha$, which incorporates also some information around the saturated regions.

The high variance, and small differences in mean between good models and the naive one, points to how this type of comparison also is inadequate, where quality of linearization still has a prominent effect on the results. There is a need for better separation of the variations in the reconstruction problem.

\begin{figure*}
    \centering
    \begin{subfigure}[b]{\textwidth}
        \centering
        \includegraphics[width=\linewidth]{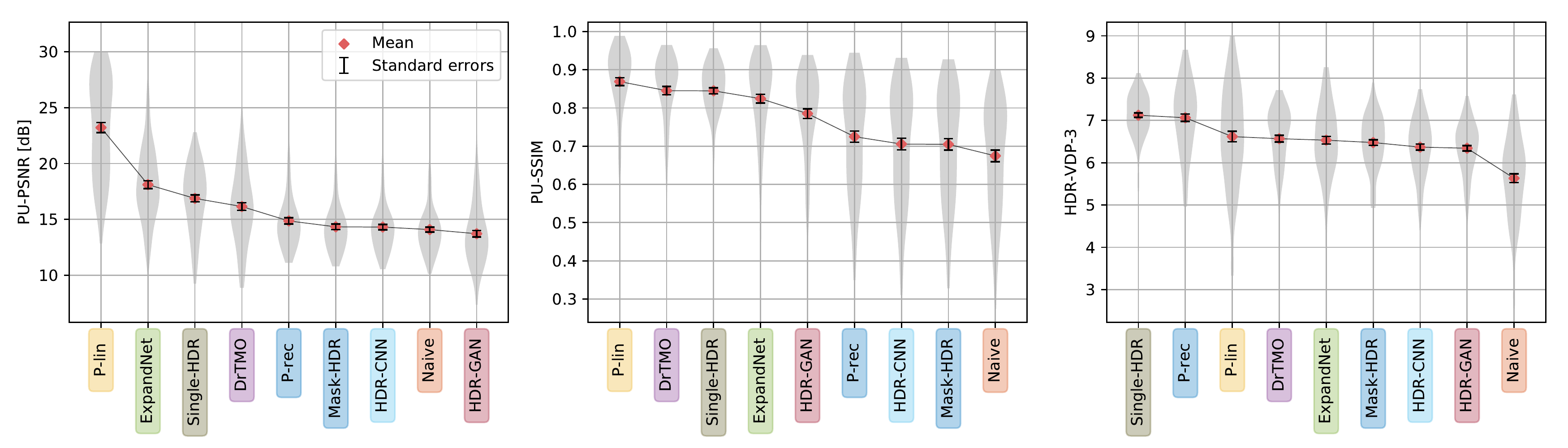}
        \caption{Camera simulation: M-CRF, EV-10}
    \end{subfigure}
    \begin{subfigure}[b]{\textwidth}
        \centering
        \includegraphics[width=\linewidth]{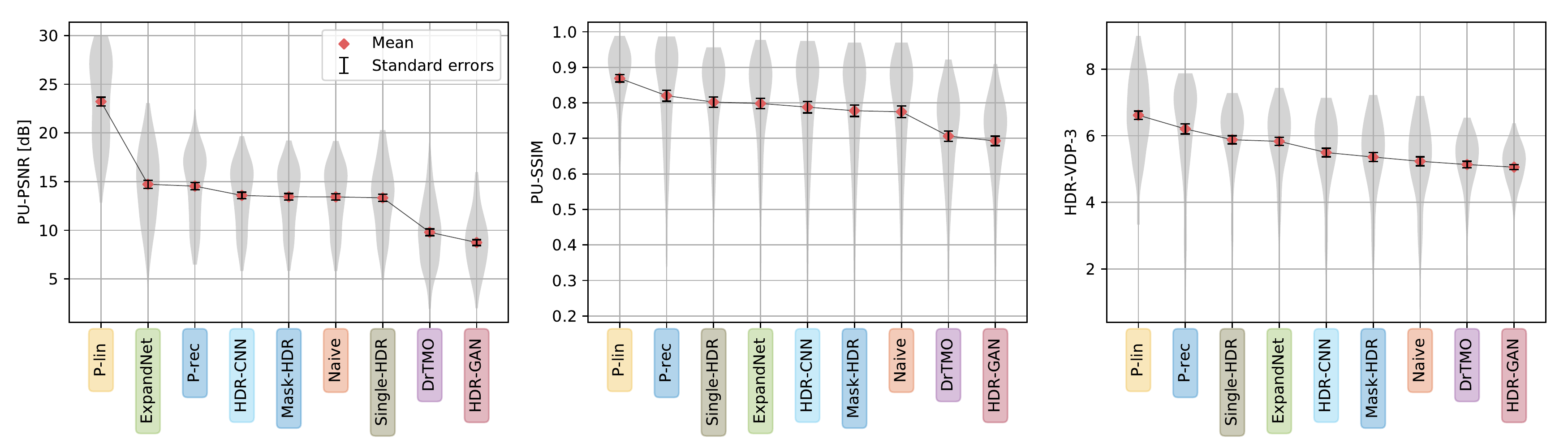}
        \caption{Camera simulation: CLAHE, EV-10}
    \end{subfigure}
    
    \caption{The distribution of metric values over the 96 tested scenes, where methods have been sorted by mean value to facilitate comparing differences in ranking. (a) uses camera simulation with M-CRF, while (b) is with CLAHE, and both have been simulated with EV-10. Left, middle, and right show results with PU-PSNR, PU-SSIM, and HDR-VDP-3, respectively.}
    \label{fig:eval}
\end{figure*}

\begin{figure*}
    \centering
    \includegraphics[width=\linewidth]{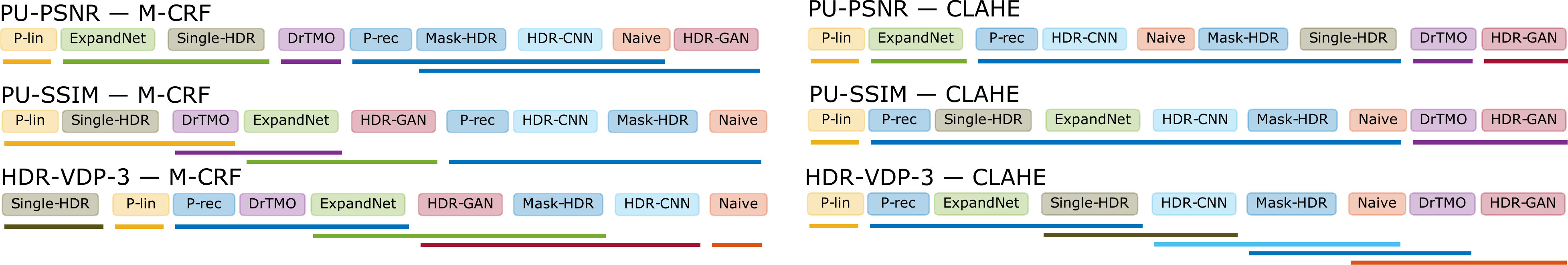}
    \caption{Rankings for different camera simulations with EV-5. The lines connect methods where the differences cannot be deemed statistically significant in a t-test, with a p-value threshold of $0.05$.}
    \label{fig:rankings5}
\end{figure*}

\begin{figure*}
    \centering
    \includegraphics[width=\linewidth]{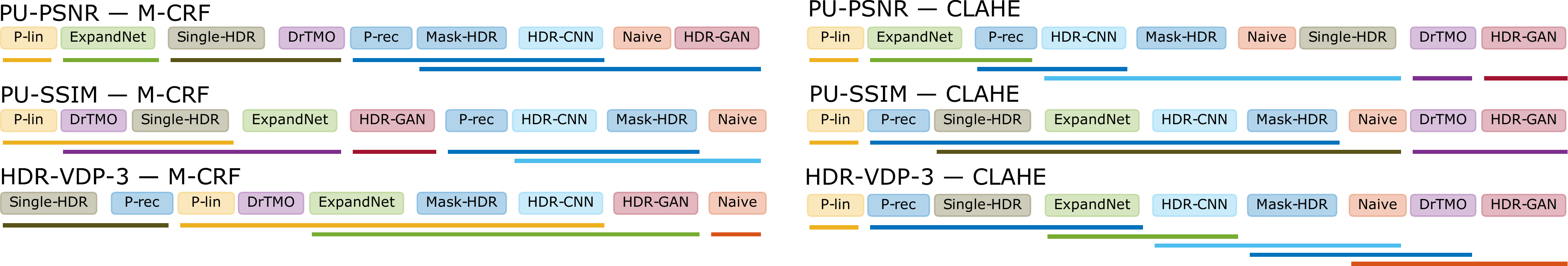}
    \caption{Rankings for different camera simulations with EV-10. The lines connect methods where the differences cannot be deemed statistically significant in a t-test, with a p-value threshold of $0.05$.}
    \label{fig:rankings10}
\end{figure*}

\begin{figure*}
    \centering
    \begin{subfigure}[b]{\textwidth}
        \centering
        \includegraphics[width=\linewidth]{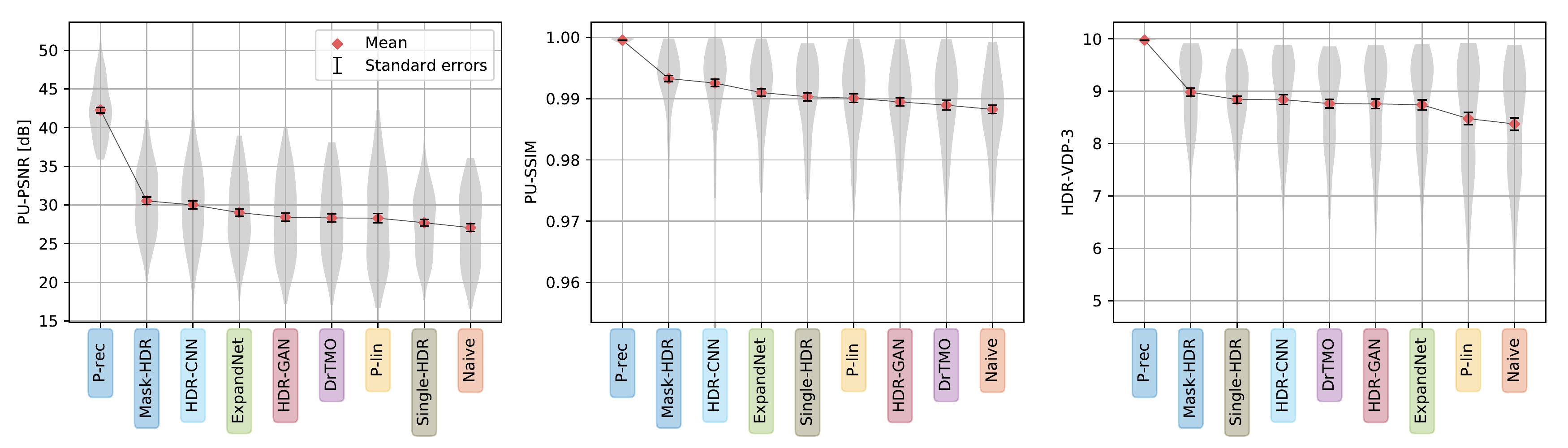}
        \caption{Camera simulation: M-CRF, EV-5}
    \end{subfigure}
    \begin{subfigure}[b]{\textwidth}
        \centering
        \includegraphics[width=\linewidth]{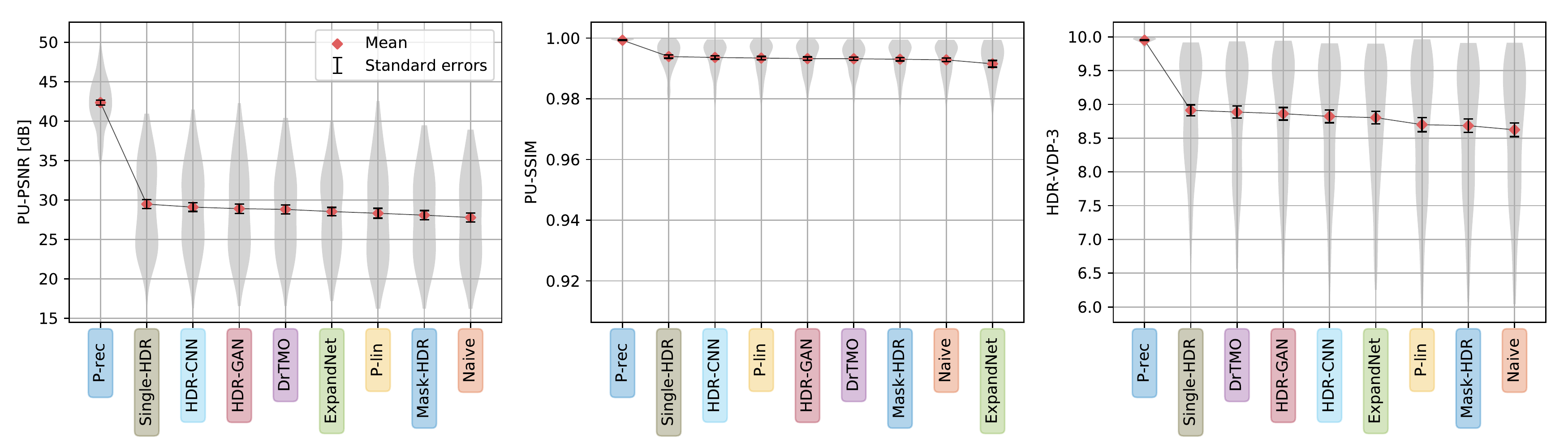}
        \caption{Camera simulation: CLAHE, EV-5}
    \end{subfigure}
    
    \caption{Same as \figref{eval}, but for EV-5 and only evaluated in saturated regions of the images.}
    \label{fig:eval_sat}
\end{figure*}

\begin{figure*}
    \centering
    \includegraphics[width=0.5\linewidth]{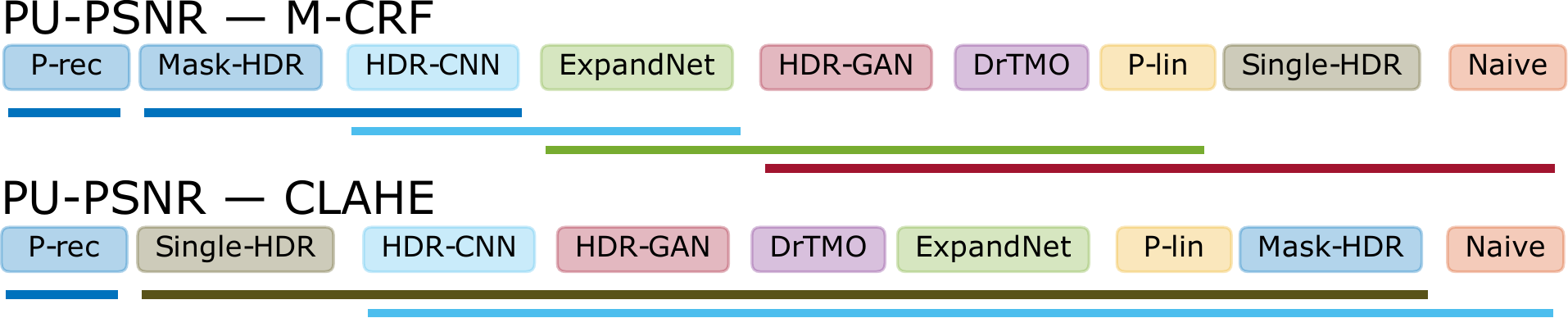}
    \caption{The rankings provided by PU-PSNR with M-CRF and CLAHE, when evaluated only in saturated image regions.}
    \label{fig:rankings_sat}
\end{figure*}

\begin{figure*}
    \centering
    \begin{subfigure}[b]{0.32\textwidth}
        \centering
        \includegraphics[width=\linewidth]{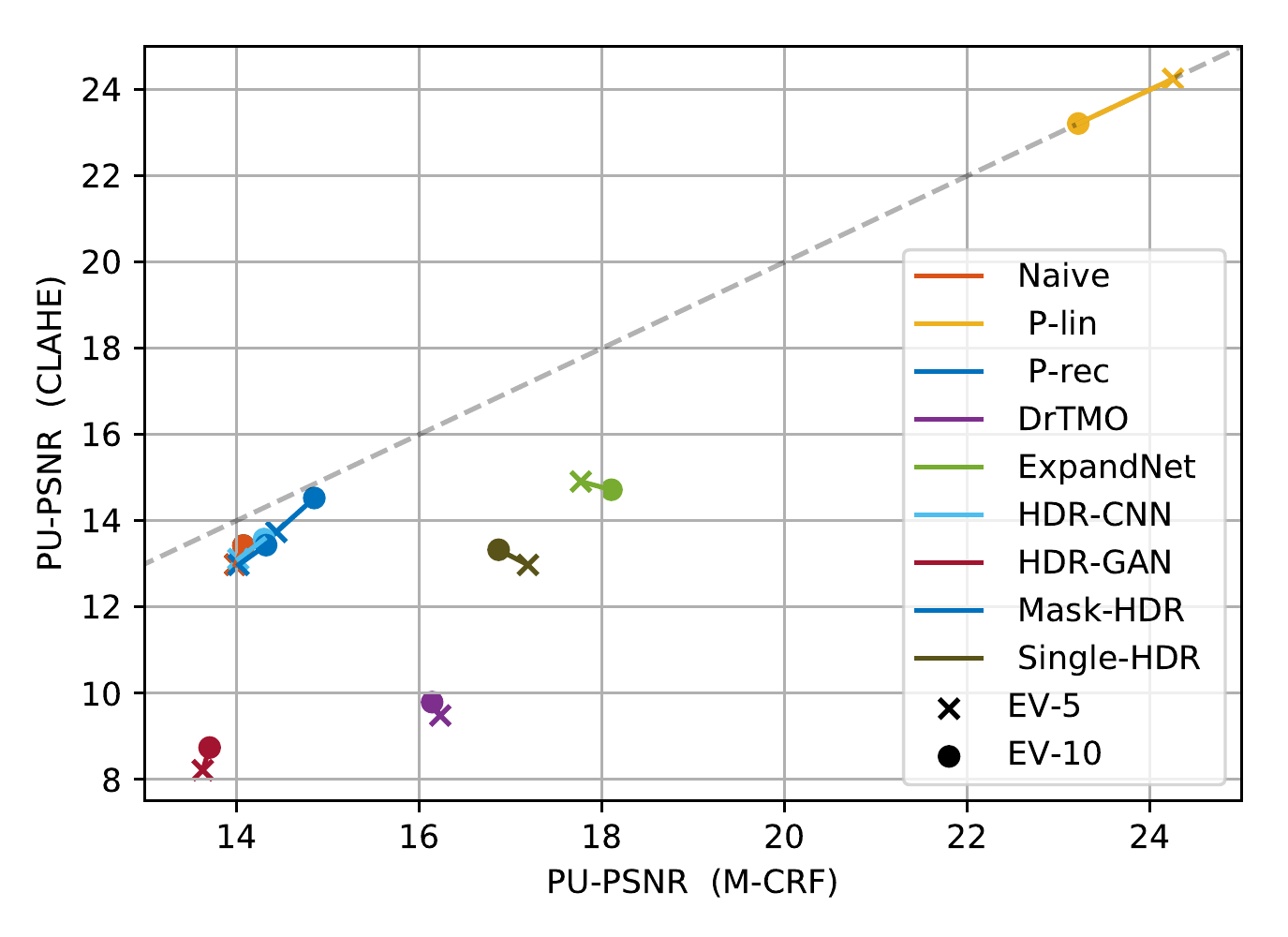}
        \caption{PU-PSNR}
    \end{subfigure}
    \begin{subfigure}[b]{0.32\textwidth}
        \centering
        \includegraphics[width=\linewidth]{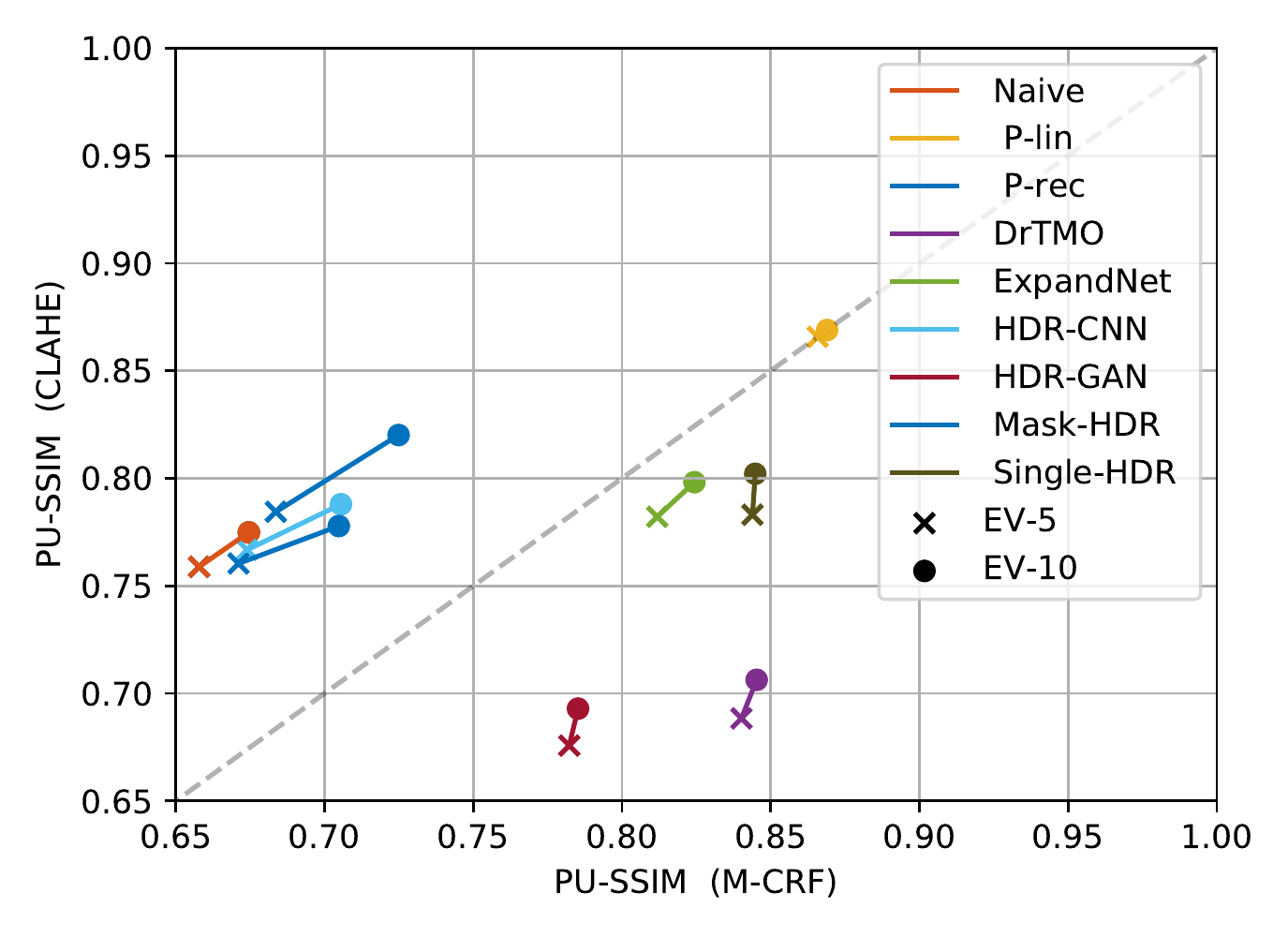}
        \caption{PU-SSIM}
    \end{subfigure}
    \begin{subfigure}[b]{0.32\textwidth}
        \centering
        \includegraphics[width=\linewidth]{fig/crf_ev.pdf}
        \caption{HDR-VDP-3}
    \end{subfigure}
    
    \caption{Differences in PU-PNSR (a), PU-SSIM (b) and HDR-VDP-3 (c) when using M-CRF and CLAHE. The two points for each method are with EV-5 and EV-10, demonstrating how many methods do not show an expected reduction in quality with increased camera exposure (more challenging reconstruction problem).}
    \label{fig:crf_ev}
\end{figure*}


\begin{figure*}
    \centering
     \includegraphics[width=\linewidth]{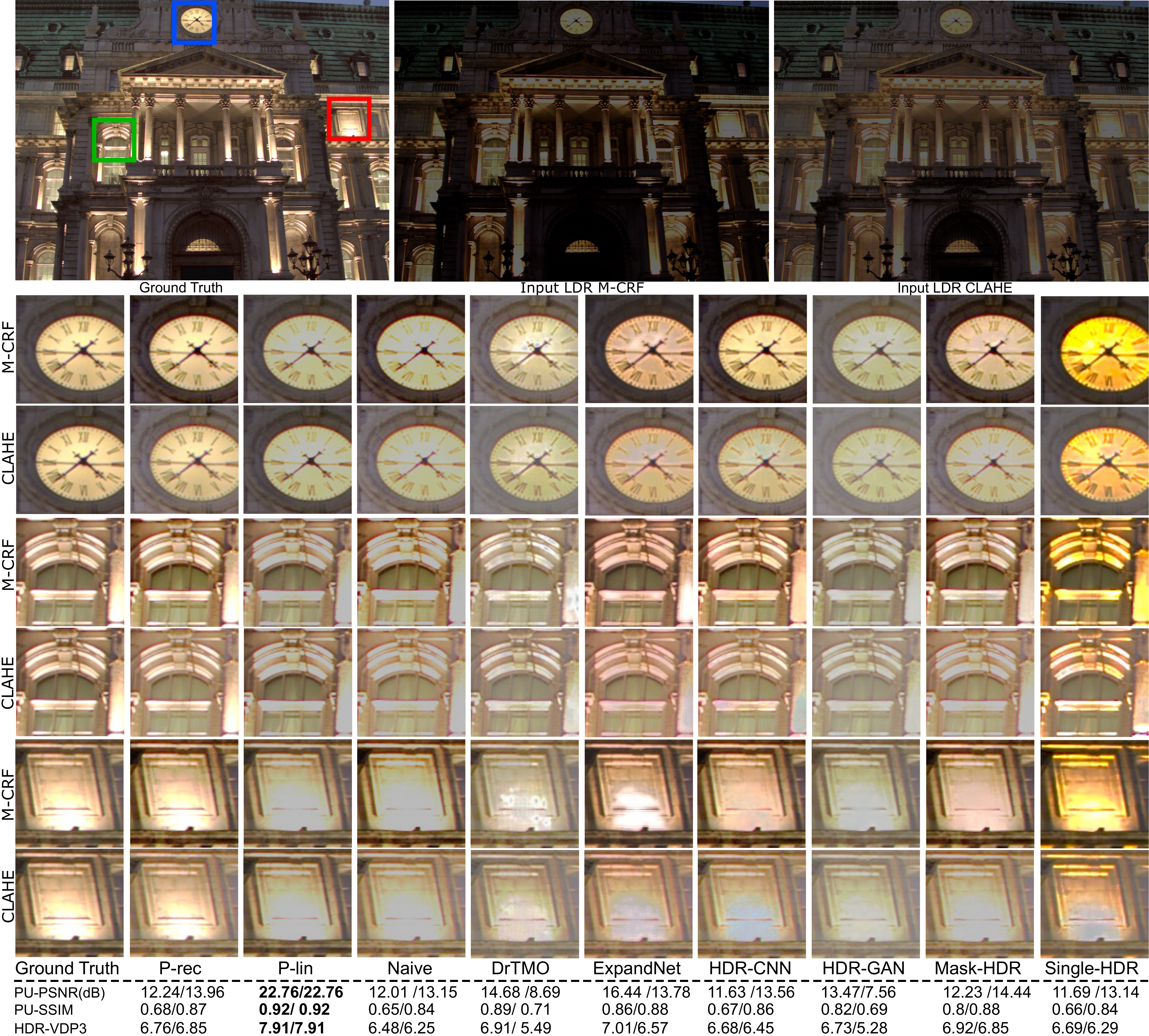}
    \caption{Selected scene areas for different reconstructions, with input LDR images simulated using M-CRF and CLAHE. The metrics in the bottom show the performance with M-CRF/CLAHE. The exposure time was set such that $5\%$ of pixels are saturated (EV-5). Ground truth and reconstructed HDR images have been gamma-encoded for display.}
    \label{fig:comparison2}
\end{figure*}

\begin{figure*}
    \centering
     \includegraphics[width=\linewidth]{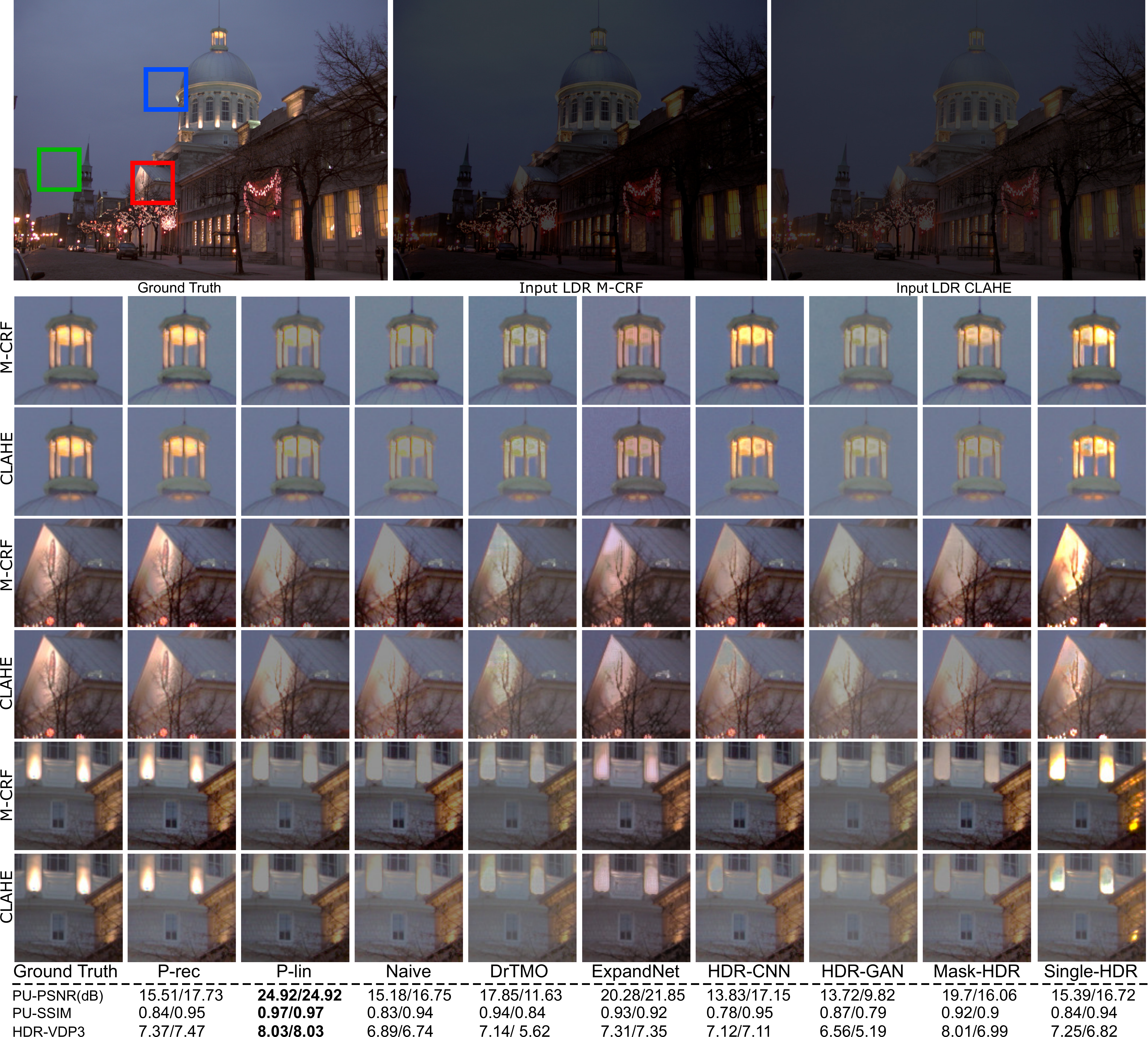}
    \caption{Selected scene areas for different reconstructions, with input LDR images simulated using M-CRF and CLAHE. The metrics in the bottom show the performance with M-CRF/CLAHE. The exposure time was set such that $5\%$ of pixels are saturated (EV-5). Ground truth and reconstructed HDR images have been gamma-encoded for display.}
    \label{fig:comparison3}
\end{figure*}

\end{document}